\pdfoutput=1

\documentclass[11pt]{article}

\usepackage[preprint]{acl}

\usepackage{times}
\usepackage{latexsym}
\usepackage{graphicx}  
\usepackage{adjustbox}
\usepackage[T1]{fontenc}

\usepackage[utf8]{inputenc}
\usepackage{newunicodechar}
\newunicodechar{⁠}{}
\usepackage{microtype}

\usepackage{inconsolata}

\usepackage{graphicx}
\usepackage{hyperref}
\usepackage{fancyvrb} 
\usepackage{tcolorbox} 
\usepackage{booktabs} 
\usepackage{caption}
\usepackage{enumitem}
\title{\textit{Grahak-Nyay:} Consumer Grievance Redressal through \\ Large Language Models}

\author{Shrey Ganatra\textsuperscript{\rm 1}, \ Swapnil Bhattacharyya\textsuperscript{\rm 1}, \ Harshvivek Kashid\textsuperscript{\rm 1}, \ Spandan Anaokar\textsuperscript{\rm 1}, \\
\textbf{Shruti Nair}\textsuperscript{\rm 2}, \ \textbf{Reshma Sekhar}\textsuperscript{\rm 2}, \ 
\textbf{Siddharth Manohar}\textsuperscript{\rm 2}, \ 
\textbf{⁠Rahul Hemrajani}\textsuperscript{\rm 2}, \\ 
\textbf{Pushpak Bhattacharyya}\textsuperscript{\rm 1} \\
\textsuperscript{\rm 1}Indian Institute of Technology Bombay\\
\textsuperscript{\rm 2}National Law School of India University, Bangalore\\
\texttt{\{ganatrashrey2002, harshvivek14, spandananao, pushpakbh\}@gmail.com}\\
\texttt{swapnilbhyya@cse.iitb.ac.in}
}

\begin{document}
\maketitle
\begin{abstract}
Access to consumer grievance redressal in India is often hindered by procedural complexity, legal jargon, and jurisdictional challenges. To address this, we present \textbf{Grahak-Nyay} (Justice-to-Consumers), a chatbot that streamlines the process using open-source Large Language Models (LLMs) and Retrieval-Augmented Generation (RAG). Grahak-Nyay simplifies legal complexities through a concise and up-to-date knowledge base. We introduce three novel datasets: \textit{GeneralQA} (general consumer law), \textit{SectoralQA} (sector-specific knowledge) and \textit{SyntheticQA} (for RAG evaluation), along with \textit{NyayChat}, a dataset of 300 annotated chatbot conversations. We also introduce \textit{Judgments} data sourced from Indian Consumer Courts to aid the chatbot in decision making and to enhance user trust. We also propose \textbf{HAB} metrics (\textbf{Helpfulness, Accuracy, Brevity}) to evaluate chatbot performance. Legal domain experts validated Grahak-Nyay's effectiveness. Code and datasets will be released.




\end{abstract}

\section{Introduction}
Large Language Models (LLMs) like GPT-4 \cite{achiam2023gpt} and Llama-3 \cite{dubey2024llama} have found widespread use in various domains, including finance \cite{zhao2024optimizing}, tourism \cite{meyer2024comparisonllmfinetuningmethods}, healthcare \cite{mishra-etal-2023-pal}, education \cite{lee-etal-2023-peep}, and customer support \cite{obadinma-etal-2022-bringing}. While LLMs have been applied to legal tasks such as judgment prediction, summarization, and case retrieval \cite{joshi-etal-2024-il,feng-etal-2024-legal}, there's a notable gap in their application to consumer law, especially for assisting individuals with everyday grievances. This is particularly crucial in India.

\begin{figure}[t]
    \centering
    \includegraphics[width=\linewidth]{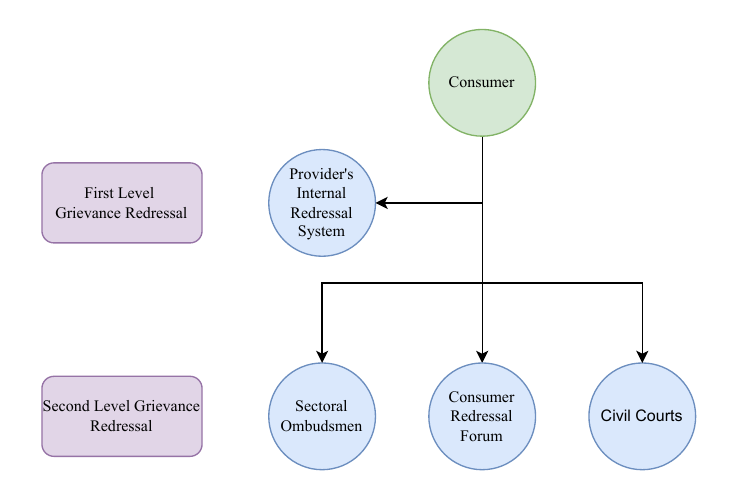}
    \caption{Two Level Grievance Redressal System in India, where the Consumer is first encouraged to approach the Provider's Internal Redressal System (like Customer Care) and then reach out to Government bodies for redressal.}
    \label{fig:two-level-system}
\end{figure}

In India, consumer grievance redressal remains a significant challenge. Despite the efforts of the Department of Consumer Affairs\footnote{\url{https://consumeraffairs.nic.in/}} through initiatives like the National Consumer Helpline\footnote{\url{https://consumerhelpline.gov.in/}} and the e-Daakhil\footnote{\url{https://edaakhil.nic.in/}} portal, many individuals without legal expertise still struggle with filing complaints. While legal representation is not required to file a consumer complaint, the complexity of legal language, jurisdictional issues, and strict limitation periods create barriers for consumers. In the financial year 2023-2024, approximately 107,966 complaints were registered every month with the National Consumer Helpline \cite{ministry-ca}, indicating a large volume of grievances that need timely attention. However, these complaints are handled by trained staff, revealing a critical gap in consumer awareness and legal literacy, and underscoring the need for tools to accelerate the grievance resolution.

Existing general-purpose chatbots like ChatGPT\footnote{\url{https://chatgpt.com/}} and Claude\footnote{\url{https://claude.ai/}} provide generic information, but lack the specific legal knowledge and interactive capabilities needed for effective consumer grievance filing (Fig. \ref{fig:chat-chatgpt}). India's two-level grievance redressal system (Fig. \ref{fig:two-level-system}) encourages direct contact with service providers, escalating to government bodies if necessary. However, complaints are frequently rejected due to issues such as as incorrect jurisdiction, misrepresentation, or failure to meet legal requirements, often stemming from a lack of understanding of consumer law \cite{rbi} (Fig. \ref{fig:complaint-rejection}).

To address this, we introduce \textbf{Grahak-Nyay} (Justice-to-Consumers), a chatbot designed to empower Indian consumers by providing the legal knowledge necessary to navigate the grievance redressal system. Grahak-Nyay assists in interpreting complex legal language, preparing documentation (complaint letters, forms) and guiding users through escalation procedures. Unlike general-purpose chatbots, Grahak-Nyay offers context-specific legal assistance, enabling informed action without requiring formal legal representation. By addressing the key challenges – lack of consumer law knowledge and documentation complexity – our chatbot aims to increase successful complaint filings and streamline the resolution process.

Our contributions are:
\begin{enumerate} 

    \item \textit{\textbf{Grahak-Nyay}}: A consumer grievance redressal chatbot tailored for Indian consumers, utilizing an open-source Large Language Model powered by Retrieval-Augmented Generation, aided by a concise Knowledge Base with the latest information (Section \ref{sec:Methodology}).
    \item \textit{\textbf{GeneralQA}}: A question-answer dataset based on general consumer laws; \textit{\textbf{SectoralQA}}: A question-answer dataset based on sector-wise knowledge of consumer laws; and \textit{\textbf{SyntheticQA}}: A question-answer dataset to evaluate the performance of RAG (Section \ref{sec:KB}).
    \item \textit{\textbf{Judgments Data}}: An annotated corpus of 570 Indian Consumer Court judgments along with summaries and categories, used to enhance user trust (Section \ref{sec:Judgments}).
    \item \textit{\textbf{NyayChat}}: A dataset containing 300 annotated conversations between users and the chatbot based on various issues and complaints. Each conversation averages 32 turns and 3,475 tokens, demonstrating the depth and richness of the interactions. (Section \ref{sec:nyaychat}).
    \item \textbf{HAB} metrics, to assess the quality of chatbot conversations based on \textbf{H}elpfulness, \textbf{A}ccuracy, and \textbf{B}revity. We conduct human-based evaluation and experiments on reference-free automatic evaluation of conversations using various \textit{LLM-based} evaluators based on the HAB metrics (Section \ref{sec:Eval}).

\end{enumerate}
\begin{table}[t]
\centering
\begin{tabular}{@{}ll@{}}
\toprule
\textbf{Statistic}                     & \textbf{Value} \\ \midrule
\textbf{Total Conversations}           & 303            \\
\textbf{Average Turns per Conversation} & 32.01          \\
\textbf{Average Tokens per Conversation} & 3475.26       \\ \bottomrule
\end{tabular}
\caption{Statistics of the NyayChat dataset, which consists of simulated conversations addressing consumer law grievances.}
\label{tab:nyaychat-key-stats}
\end{table}

\begin{figure*}
\centering
    \tcbset{colframe=blue,colback=blue!10, boxrule=0.5mm, arc=0mm}
    \begin{tcolorbox}
    \small
User: Builder is not giving possession, any past judgments that can help me, please cite them?
\\
\\
Grahak-Nyay: I'd be happy to help you with that.
In a case where possession was offered after a long delay, but the complainants refused to take possession as they no more needed the flat after such delay, it was held that only the earnest money/initial deposit paid by the buyer can be forfeited by the builder in case the buyer does not want to continue in the project due to any default of the builder. \textbf{(CC/438/2019/2020, NCDRC)}

Additionally, in a case where the builder delayed handing over flats after collecting a significant amount of money, the National Consumer Disputes Redressal Commission (NCDRC) held that the builder is liable for deficiency in service and must compensate the buyers for the delay. \textbf{(FA/347 of 2021/2024, NCDRC)}

In another case, where possession was delayed for nearly 7 years, the NCDRC held that the delay in possession was a deficiency of service, justifying compensation. \textbf{(CC/197/2013/2022 NCDRC)}

These judgments may be relevant to your case, but please note that each case is unique, and the outcome may vary depending on the specific circumstances.

Would you like to know more about the process of filing a complaint against the builder or drafting a legal notice?
    \end{tcolorbox}
    \caption{Chat example from Grahak Nyay using the Judgments Database to answer user queries}
    \label{fig:chat-judgment}
\end{figure*} 
\section{Dataset}
\label{sec:Dataset}
We introduce four datasets: three forming the knowledge base for Retrieval-Augmented Generation (RAG), and one containing simulated user-chatbot conversations. For more detail on the dataset curation process, refer to Section \ref{sec:datasetCuration}.

\subsection{Knowledge Base}
\label{sec:KB}
\subsubsection{GeneralQA}
This dataset provides a structured overview of the Consumer Protection Act 2019 \cite{cpa}, consisting of 53 question-answer pairs. It covers the Act's key provisions and defines fundamental concepts like 'consumer' and 'goods,' clarifying the chatbot's operational scope.  This dataset establishes the foundational legal framework for the chatbot.

\subsubsection{SectoralQA}
This dataset encompasses detailed information from 23 sectors, including real estate, telecommunications, sports and recreation. The corpus comprises 889 questions and answers, each tailored to address sector-specific concerns. A breakdown of the sector-wise distribution, along with the general corpus, is provided in Fig. \ref{fig:distribution}.

\subsubsection{SyntheticQA}
To assess our RAG system, we generated five user questions that yield the same answer across both the general corpus and the sector-specific corpus, using the Llama-3.1-70B-Instruct\footnote{\url{https://huggingface.co/meta-llama/Llama-3.1-70B-Instruct}} model. This amounts to generating 4734 (in some cases, 6 instead of 5 questions) questions. The answers were generated using the Llama-3.1-8B-Instruct\footnote{\url{https://huggingface.co/meta-llama/Llama-3.1-8B-Instruct}} model, keeping the corresponding answer in the context. We call this dataset as \textbf{SyntheticQA}.
The prompt used for generation can be found in Fig.\ref{fig:SyntheticQA prompt}.

\subsection{Judgments}
\label{sec:Judgments}
We introduce a novel, expert-annotated corpus of judgments from Consumer Courts in India, a critical resource for advancing NLP research in the legal domain. This corpus comprises 570 judgments, meticulously curated and spanning 23 distinct sectors of Indian consumer law. Each judgment has been summarized and categorized by legal experts, ensuring high-quality annotations and domain-specific relevance.

To the best of our knowledge, this represents the first publicly available dataset of its kind, addressing a significant gap in resources for legal NLP, particularly within the Indian judicial context. The dataset's immediate utility is demonstrated in its application within our legal chatbot (Figure \ref{fig:chat-judgment}), where it enables informed decision-making and allows for the citation of pertinent case law to enhance user trust and transparency.

Beyond this specific application, the corpus holds substantial potential for a range of NLP tasks. Its rich annotations and structured nature make it highly suitable for benchmarking and developing models for legal text summarization, case classification, legal information retrieval, and potentially for predicting case outcomes or identifying relevant legal precedents. We believe this dataset will be an invaluable asset to the NLP community, fostering further research and development in the underexplored intersection of artificial intelligence and consumer law.



\begin{figure*}
    \centering
    \includegraphics[width=0.7\linewidth]{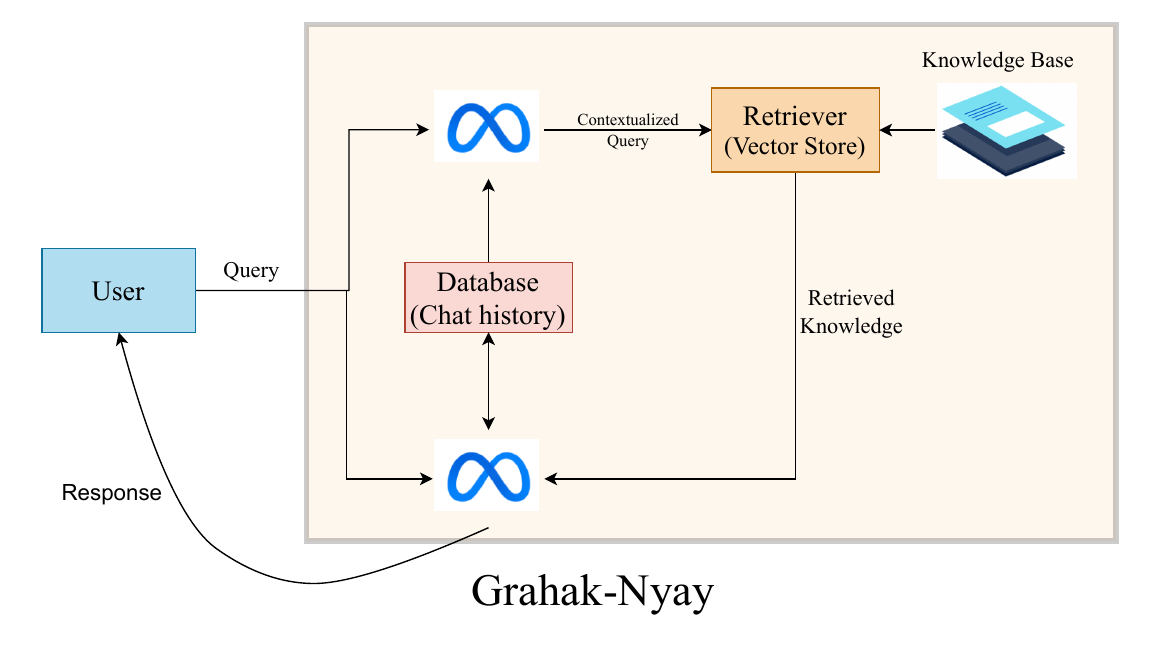}
    \caption{Architecture of \textit{Grahak-Nyay} chatbot. The user query is first contextualised using an LLM to form an independent query to perform RAG. The retrieved knowledge is then given to the LLM along with the chat history to generate an appropriate response to the user query.}
    \label{fig:architecture}
\end{figure*} 
\subsection{NyayChat}
\label{sec:nyaychat}
This dataset includes 303 simulated conversations meticulously crafted by a team of legal experts specializing in various sectors such as e-commerce, medical negligence, railways, airlines, and more. Each conversation mirrors a real-world interaction between a user and the chatbot, addressing specific grievances that fall under the purview of consumer law. Each conversation averages 32.01 turns and 3,475.26 tokens 
(Tab. \ref{tab:nyaychat-key-stats}), demonstrating the depth and richness of the interactions. This dataset serves as a valuable benchmark for advancing research in conversational AI, particularly in the domain of user grievance redressal. 

\begin{figure*}[t]
    \centering
    \small
    \includegraphics[width=\linewidth]{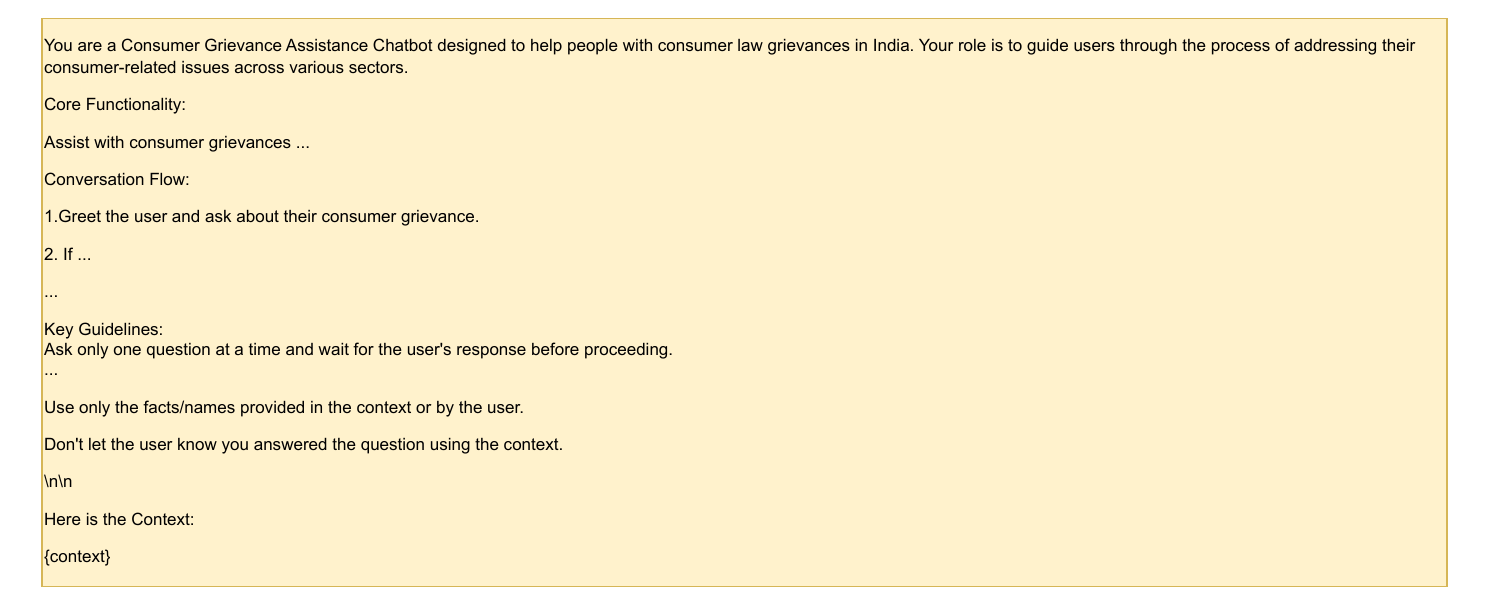}
    \caption{Part of system prompt designed for Grahak-Nyay Chatbot with a structured conversation flow: it gathers grievance details step-by-step, offers remedies under Indian consumer law, assists in drafting legal documents (e.g., notices, complaints), guides users on using the National Consumer Helpline and e-daakhil portal, and provides tailored responses while strictly limiting interactions to consumer-related issues. For entire prompt see Figure \ref{fig:system-prompt-long}}.
    \label{fig:system prompt}
\end{figure*}
\section{Methodology}
\label{sec:Methodology}

\subsection{Retrieval-Augmented Generation}
Retrieval Augmented Generation (RAG) is a prominent approach used in real-world applications for grounding large language model (LLM) generations in up-to-date and domain-specific knowledge. It has been observed \cite{lazaridou2022internetaugmentedlanguagemodelsfewshot, shuster2021retrievalaugmentationreduceshallucination, ren-etal-2023-retrieve} that RAG reduces hallucinations and improves answer quality, without the need for
highly expensive and sometimes fragile domain-specific
fine-tuning. 

A typical RAG framework involves a retrieval system that fetches documents that are relevant to the query. These documents are then used as context, prompting the LLM to generate the required response. For our chatbot, we also include the chat history as part of the prompt. The RAG framework is divided into the following parts in our architecture.

\subsubsection{Chunking}
We observed that fixed-length chunking proved inadequate for our purposes. Long answers were often split across multiple chunks, resulting in the loss of crucial information during retrieval. Conversely, grouping multiple short answers in a single chunk introduced unnecessary noise. As a result, we adopted a chunking strategy where each chunk contains only one Question-Answer pair, ensuring clarity and precision in information retrieval.

\subsubsection{Query Rewriting}

To implement RAG, it is essential for each query to be properly contextualized. To accomplish this, we utilized Llama-3.1-8B-Instruct along with the instruction provided in Fig.\ref{fig:query-rewriting-prompt}. Initially, we observed that the LLM answered the query directly rather than rewriting it as intended. To improve performance, we utilized one-shot prompting by incorporating a single example in the prompt, which led to significantly better results.

\begin{table*}[h]
\centering
\scriptsize
\begin{tabular}{*{6}{p{1.25cm}}*{3}{p{1.25cm}}}
\toprule
\multicolumn{6}{c}{\textbf{Reference-based evaluation}} & \multicolumn{3}{c}{\textbf{Reference-free evaluation}} \\
\cmidrule(lr){1-6}\cmidrule(lr){7-9}
\textbf{ROUGE-1} & \textbf{ROUGE-2} & \textbf{ROUGE-L} & \textbf{BERTScore} & \textbf{METEOR}   & \textbf{BLEU} & \textbf{Helpfulness} & \textbf{Accuracy} & \textbf{Brevity} \\
\midrule
66.9 & 41.1 & 33.2 & 90.9 & 41.9 & 37.4 & 4.65 & 3.61 & 3.12 \\
\bottomrule
\end{tabular}
\caption{Performance of \textit{Grahak-Nyay chatbot} on Reference-based and Reference-free evaluation. We evaluated the Grahak-Nyay chatbot on 65 conversations for which reference was available. We performed LLM-based automatic evaluation on HAB metrics on the 5-point Likert scale using the gpt-4o-mini model.}
\label{tab:grahak_evaluation}
\end{table*}
\subsubsection{Retriever}
The chunks are embedded using mixedbread-ai/mxbai-embed-large-v1\footnote{\url{https://huggingface.co/mixedbread-ai/mxbai-embed-large-v1}} and stored in a vectorstore.
The retriever is responsible for extracting relevant chunks using the query. We use cosine similarity to extract the best four chunks and use it for response generation.
\subsubsection{Generation}
The chatbot is meant to be interactive and conversational. Hence, for each user input, while keeping the retrieved documents as a reference, the chatbot also takes into account the chat history. This ensures that all responses are relevant and grounded in the contextual history of the issue. We use Llama-3.1-8B-Instruct model to generate our responses.

\subsection{Hallucination}
A significant barrier to the wide use of LLMs in multiple domains is their tendency to hallucinate. It has been observed that in spite of clear instructions, LLM generates text which might be false or irrelevant. RAG-based approaches help to tremendously reduce this phenomenon, but even then, hallucination remains a major concern.

A domain-specific chatbot like our Grahak-Nyay chatbot implies that RAG is responsible for providing much of the context and domain knowledge that will be utilized for the chat. In such cases, the LLM must generate content based on the RAG Corpus only and never contradict it. For this purpose, we modified the prompt where we specified the LLM to answer any out-of-corpus question by stating that it does not know the answer.

\begin{figure}[t]
    \centering
\includegraphics[width=0.5\textwidth,height=50mm]{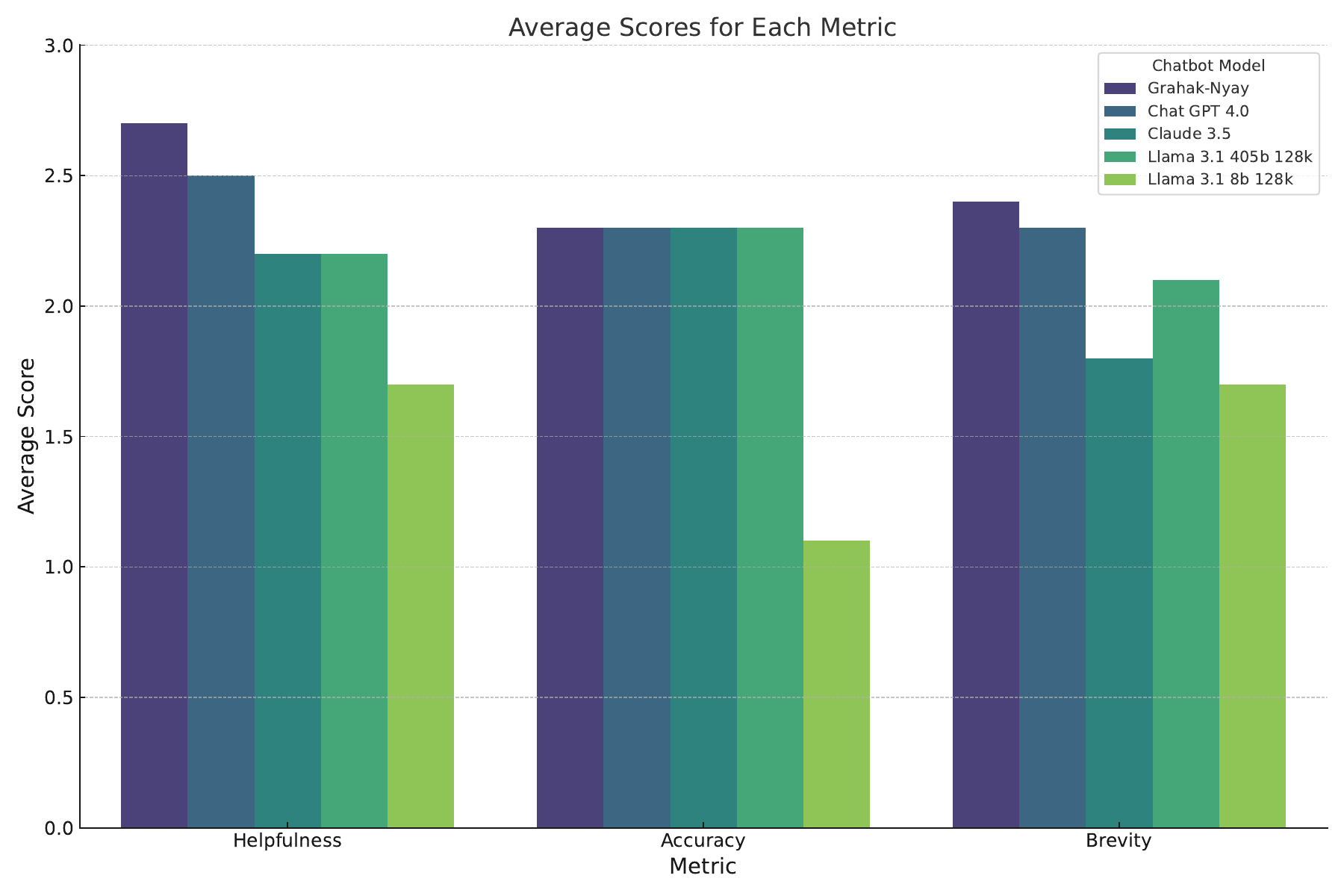}
\caption{Benchmarking of \textit{Grahak-Nyay} against other chatbots on HAB metrics by human legal experts. Grahak-Nyay outperforms in Helpfulness and Brevity. It performs similar to other larger models while outperforming the base model in terms of Accuracy.}
\label{fig:comparison}
\end{figure}


\begin{table*}[!htpb]
\centering
\small
\label{tab:bleu_rouge}
\begin{tabular}{l|cccccccc}
\toprule
\textbf{Dataset} & \textbf{BLEU} & \textbf{ROUGE-1} & \textbf{ROUGE-L} & \textbf{BERTScore} & \textbf{Ans-Rel.} & \textbf{Faithfulness} \\
 \midrule
SectoralQA   & 49.38  & 64.20  & 60.39    & 90.94  & 7.44 & 8.58 \\
GeneralQA   & 49.45  & 66.66  & 63.74   & 95.18  & 7.35 & 9.02 \\
SyntheticQA    & 31.04  & 48.37  & 40.44  & 87.93 & 7.48 & 9.30  \\
\bottomrule
\end{tabular}
\caption{Performance on BLEU, ROUGE, and BERTScore, along with automatic evaluation using RAGAS assessment based on Answer Relevance (Ans-Rel.) and Faithfulness metrics across three datasets.}
\label{tab:bleu}
\end{table*}



\section{Evaluation}
\label{sec:Eval}
We strongly believe that any user-facing chatbot should help the user address the query, be accurate while doing so, and keep the user engaged. We assess the quality of chatbot conversations using \textbf{HAB metrics}: \textbf{H}elpfulness, \textbf{A}ccuracy, and \textbf{B}revity. HAB metrics allow us to assess how effectively the chatbot addresses user issues and provides accurate information and how concisely it communicates these responses. We also qualitatively assess the chatbot performance (Appendix \ref{fig:chat1}, \ref{fig:chat2}, \ref{fig:chat3} and \ref{fig:chat4}) where multiturn conversations of human followed by chatbot has been presented.
\\
The HAB metrics are defined as follows:
\begin{itemize}[leftmargin=5pt]
    \item \textbf{Helpfulness}: This metric assesses how helpful the chatbot was in resolving the user's issue or query. It evaluates the chatbot's ability to understand the user's problem and provide actionable, relevant, and clear resolution.
    \item \textbf{Accuracy}: This metric evaluates the correctness of the information provided by the chatbot in response to user queries, ensuring that the responses are factually accurate and reliable.
    \item \textbf{Brevity}: This metric measures the conciseness of the chatbot's responses, ensuring efficient communication without unnecessary elaboration. It ensures efficient communication by focusing on delivering essential information straight to the point while avoiding excessive questioning or verbosity.
\end{itemize}
\subsection{Human Evaluation of other chatbots}
Using the HAB metric, we evaluated publicly available chatbots, including \texttt{ChatGPT-4.0}, \texttt{Claude-3.5}, \texttt{Llama-3.1-405b-128k}, and \texttt{Llama-3.1-8b-128k}, with assessments conducted by human \textit{legal} experts on 5-point Likert scale. The analysis revealed that the \textit{Grahak-Nyay} chatbot surpassed all other chatbots on the HAB metrics (Fig. \ref{fig:comparison}).

\subsection{Evaluation of Grahak-Nyay chatbot}
We present the detailed results from the evaluation of 65 chats obtained using the Grahak-Nyay chatbot in Table \ref{tab:grahak_evaluation}, categorized into two groups: Reference-based and Reference-free evaluations. For these 65 chats, reference responses annotated by the legal experts were available, enabling the application of Reference-based metrics. Additionally, for the Reference-free evaluation, we utilized HAB metrics to assess the chatbot's performance in providing relevant and concise responses. We used the best performing model, gpt-4o-mini model, which demonstrated the highest correlation with human evaluations, for the LLM-based assessment of the HAB metrics on 5-point Likert scale (Appendix \ref{sec:auto_eval}).

\subsection{Evaluation of Retrieval-Augmented Generation}
We evaluate our Retrieval-Augmented Generation (RAG) system by asking questions from GeneralQA, SectoralQA, and SyntheticQA. The system is assessed using BLEU, ROUGE, and BERTScore, along with automatic evaluation metrics such as Faithfulness and Answer Relevance using \textbf{RAGAS}~\cite{es2023ragasautomatedevaluationretrieval}. Detailed results are presented in Table~\ref{tab:bleu}.  Faithfulness measures whether the generated answer is grounded in the provided context. Answer Relevance evaluates how well the generated answer addresses the given question. To assess relevance, we compare the generated response with the ground truth. We use the gpt-4o-mini model as an evaluator. Faithfulness and Answer Relevance are scored on a 0-10 scale, while other metrics are measured on a 0-100 scale.
\section{Deployment}
We utilize the Text Generation Inference (TGI)\footnote{\url{https://huggingface.co/docs/text-generation-inference/en/index}} toolkit (v3.2.1) to serve the \texttt{Llama-3.1-8B-Instruct} model. TGI provides a production-ready server with features crucial for real-world deployment, including continuous batching of incoming requests for increased throughput, prefix caching to reduce redundant computations, and token streaming using Server-Sent Events (SSE) for a responsive user experience.

Our current deployment utilizes a single NVIDIA A100 GPU with 40GB of memory through its official docker image (Fig. \ref{fig:script-tgi}).

We plan on incorporating auto-scaling and adding high availability to handle potential outages.

\section{Conclusions}
In this work, we introduced our \textit{Grahak-Nyay} chatbot to address consumer grievances in various sectors. We evaluated the chatbot performance using traditional NLP metrics, automated evaluation by LLMs, and human evaluation by legal experts. Using a RAG-based framework and prompts designed to prevent hallucinations, the chatbot demonstrated the ability to handle consumer grievances in an approachable and informative way. The chatbot presents an opportunity for many people who are hesitant to take action on their consumer complaints due to a lack of complete information and help them get justice. 


\bibliography{acl}
\newpage

\appendix
\section{Appendix}
\label{sec:appendix}

\begin{table*}[ht]
\centering
\small 
\begin{tabular}{
  c  
  *{2}{p{1.0cm}}  
  *{2}{p{1.0cm}}  
  *{2}{p{1.0cm}}  
}
\toprule
 & \multicolumn{2}{c}{\textbf{Helpfulness}} & \multicolumn{2}{c}{\textbf{Accuracy}} & \multicolumn{2}{c}{\textbf{Brevity}} \\
\cmidrule(lr){2-3}\cmidrule(lr){4-5}\cmidrule(lr){6-7}
\textbf{Models} & \textbf{$r_{pb}$} & \textbf{$\rho$} & \textbf{$r_{pb}$} & \textbf{$\rho$} & \textbf{$r_{pb}$} & \textbf{$\rho$} \\
\midrule
Gemma-2-9B & 0.256 & 0.242 & 0.113 & 0.102 & 0.183 & 0.182 \\ [0.1cm]
Llama-3.1-8B & 0.386 & 0.246 & 0.225 & 0.213 & 0.154 & 0.153 \\[0.1cm]
Mixtral-8x7B & 0.557 & 0.490 & 0.205 & 0.207 & 0.159 & 0.141 \\[0.1cm]
Llama-3.1-70B & \underline{0.689} & \underline{0.627} & \textbf{0.461} & \underline{0.430} & \underline{0.430} & \underline{0.418} \\[0.1cm]

gpt-4o-mini & \textbf{0.719} & \textbf{0.687} & \underline{0.459} & \textbf{0.465} & \textbf{0.473} & \textbf{0.435} \\[0.1cm]
\bottomrule
\end{tabular}
\caption{Performance metrics for various models based on Helpfulness, Accuracy, and Brevity metrics. Each metric includes point biserial correlation ($r_{pb}$) and Spearman's rank correlation coefficient ($\rho$) scores for each model. The best scores are bolded, and the second-best scores are underlined.}
\label{tab:model_comparison}
\end{table*}

\subsection{Automated Evaluation of Grahak-Nyay chatbot using LLMs}
\label{sec:auto_eval}
To reduce human effort in evaluating the chatbot according to HAB metrics, we employ LLM-based automatic evaluation. The LLM evaluators are instructed to assign scores on a 5-point Likert scale and provide detailed explanations for their assigned scores using the structured prompt (Fig. \ref{fig:hab-helpfulness}, \ref{fig:hab-accuracy}, and \ref{fig:hab-brevity}). The prompt includes task description, scoring instructions based on the HAB metrics, as well as the conversation which is to be evaluated and the context\footnote{Context is passed only for the \textit{Accuracy} metric.}.

We evaluated 75 conversations for which we have human-evaluated data available in binary form (Yes, if the metric is followed, No if not), on the HAB metrics, using different LLMs sourced from HuggingFace\footnote{\url{https://huggingface.co}} and Groq\footnote{\url{https://groq.com/}}. The table \ref{tab:model_comparison} summarizes the performance of LLM-based evaluators for HAB metrics. We applied point biserial correlation to assess the relationship between the available binary human evaluation and the ordinal LLM scores from the 5-point Likert scale. This correlation is particularly useful in determining how well the LLM evaluations align with the binary outcomes. Additionally, we used Spearman correlation to evaluate the rank order of scores, providing further insights into the agreement between human and LLM evaluations. The Llama-3.1-70B model outperformed other open-source models across all three metrics, and gpt-4o-mini achieved the highest point biserial correlation and Spearman's correlation coefficients with \textit{p-value<0.05}, indicating its superior effectiveness.

\begin{figure}[t]
    \centering
    \includegraphics[width=\linewidth]{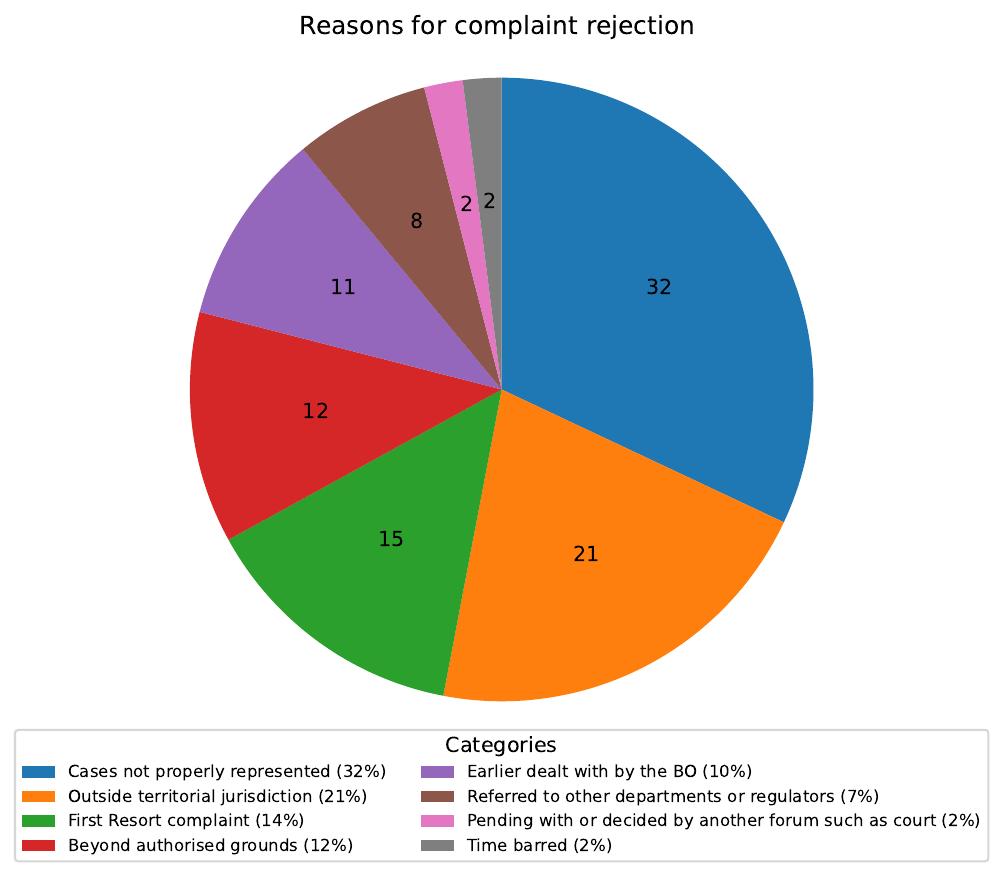}
    \caption{Reasons for Banking Ombudsman rejecting complaints in FY 2018-19. Cases not properly represented are the major reason for the rejection of complaints, followed by outside proper jurisdiction.}
    \label{fig:complaint-rejection}
\end{figure}
\begin{figure}
    \centering
    \includegraphics[width= \linewidth]{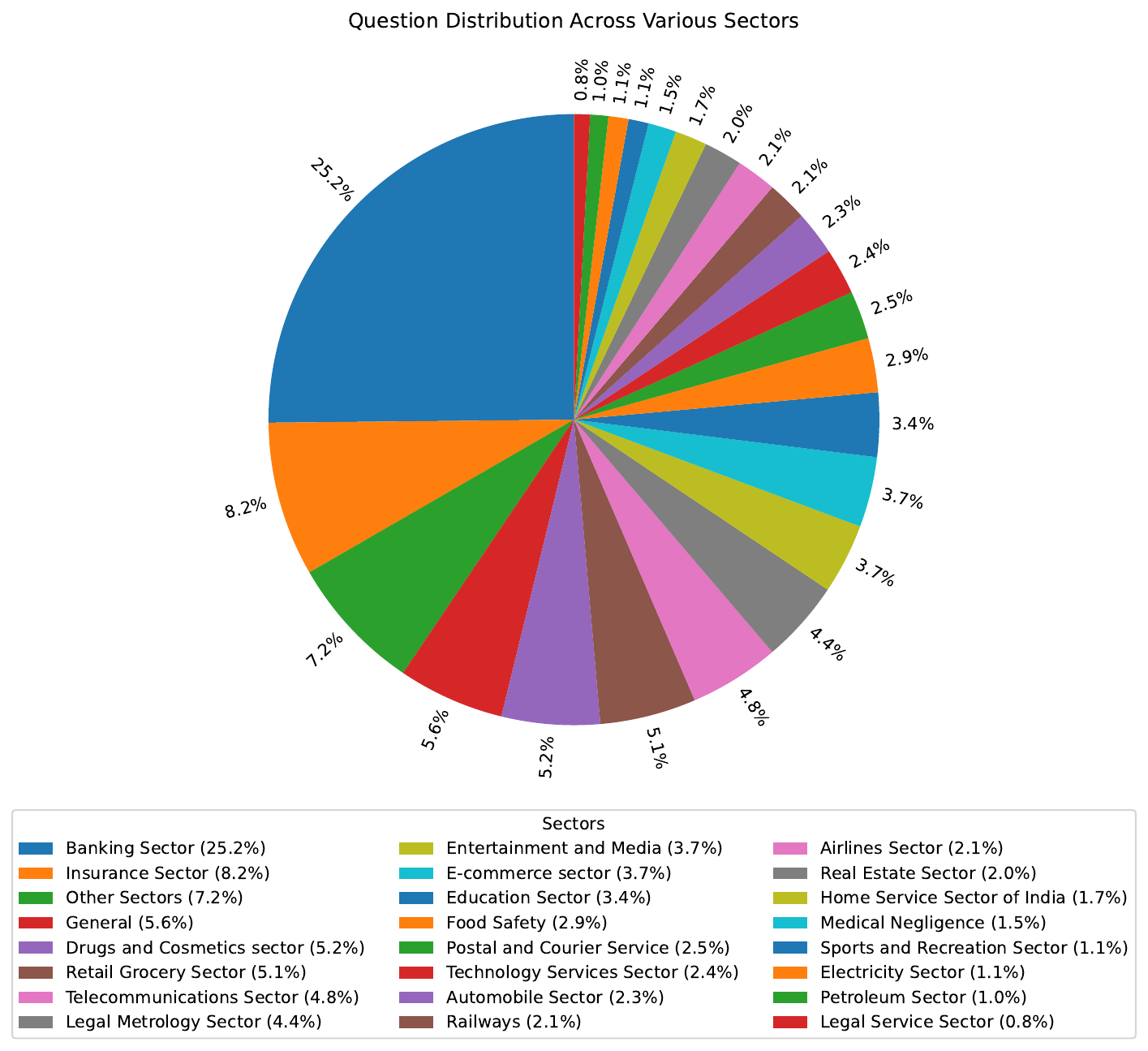}
    \caption{Distribution of corpus questions across different consumer sectors}
    \label{fig:distribution}
\end{figure}
\subsection{Human evaluation of Grahak-Nyay chatbot}

To assess the performance of our chatbot and benchmark it against several other systems, we conducted a human evaluation of the chatbot dialogues based on the HAB metrics as outlined in \ref{fig:comparison}. This evaluation was performed by a group of legal experts from the XX who were provided with a predefined set of evaluation criteria.

The evaluation of each conversation was conducted using the following rubric:

\subsubsection{Helpfulness}

\textbf{Statement:} The chatbot delivers meaningful assistance that contributes to resolving the user's issue.

\begin{itemize}
\setlength{\itemsep}{0.2em}
    \item \textbf{Score 5} - Strongly Agree: The chatbot fully addressed the issue or provided explicit, actionable steps for resolution.
    \item \textbf{Score 4} - Agree: The chatbot resolved the issue to a large extent, though minor additional guidance was required.
    \item \textbf{Score 3} - Neutral: The chatbot provided some assistance, but the response was insufficient to fully resolve the issue.
    \item \textbf{Score 2} - Disagree: The chatbot's assistance was incomplete and omitted key information.
    \item \textbf{Score 1} - Strongly Disagree: The chatbot's response was irrelevant or ineffective in resolving the issue.
\end{itemize}

\subsubsection{Accuracy}
\textbf{Statement:} The chatbot provides precise and reliable information, including correct references such as websites, phone numbers, and legal details.

\begin{itemize}
    \item \textbf{Score 5} - Strongly Agree: All information provided is entirely accurate and contextually appropriate.
    \item \textbf{Score 4} - Agree: Most information provided is accurate, with only minor, non-critical inaccuracies.
    \item \textbf{Score 3}: Neutral - The chatbot provided accurate information, but there were notable factual errors.
    \item \textbf{Score 2}: Disagree - The response contained multiple inaccuracies that could mislead the user.
    \item \textbf{Score 1}: Strongly Disagree - The information provided was largely or completely incorrect and misleading.
\end{itemize}

\subsubsection{Brevity}
\textbf{Statement:} The chatbot communicates efficiently, offering clear and concise responses without superfluous information or unnecessary queries.

\begin{itemize}
    \item \textbf{Score 5} - Strongly Agree: The response was succinct and included only the essential information.
    \item \textbf{Score 4} - Agree: The response was mostly concise, with minor extraneous details.
    \item \textbf{Score 3} - Neutral: The response included some unnecessary details or questions, reducing conciseness.
    \item \textbf{Score 2} - Disagree: The response was overloaded with irrelevant or redundant information, causing potential confusion.
    \item \textbf{Score 1} - Strongly Disagree: The response was excessively long and contained irrelevant or unnecessary information.
\end{itemize}

\subsubsection{Evaluation Procedure}
The evaluation of the Grahak-Nyay chatbot was conducted by two independent experts, using a blind evaluation methodology. The experts assessed 65 chats that were utilized in the automatic evaluation phase and assigned scores to each chat in terms of the HAB metrics. The evaluation of the remaining four chatbots, namely \texttt{ChatGPT-4.0}, \texttt{Claude-3.5}, \texttt{Llama-3.1-405b-128k}, and \texttt{Llama-3.1-8b-128k}, was conducted by three legal experts, following a similar blind methodology, using a representative subset of dialogues from each chatbot.

\subsection{Dataset Details and Curation Process}
\label{sec:datasetCuration}

\subsubsection{GeneralQA and SectoralQA}

The Knowledge Base consists of two core parts. The first is a GeneralQA on Consumer Protection in India, synthesizing general consumer grievance information into 52  question-answer pairs. These Q\&A pairs span a wide range of consumer protection topics, including the Consumer Protection Act of 2019, definitions of a “consumer”, details on filing consumer complaints, and overviews of online and offline redressal avenues. It also contains contact information for National, State, and District Consumer Dispute Redressal Commissions. By filtering out overly technical content, the focus remains on practical guidance: what counts as a consumer grievance, how to initiate legal action, and strategic advice on whether to send a notice or file directly in a consumer forum.

Alongside the GeneralQA, the SectoralQA includes 27 sector-specific documents, each addressing a specific consumer-related domain such as Banking, Telecom, or Insurance. Each of these contains around 30 question-answer pairs, with the total corpus having 835 Q\&A pairs.

\subsubsection{NyayChat}
In addition to the Knowledge Base, NyayChat was developed to support detailed examples and evaluations. This dataset provides curated, real-world scenarios and user queries, allowing the LLM to be rigorously trained and tested on realistic problem statements. The dataset consists of synthetic conversations that simulate ideal interactions between the Grahak-Nyay and users seeking remedies under consumer protection laws. These synthetic chats were drafted by law students and underwent review by two legal experts. The starting point for each chat was a real-world case sourced from a database of 1,200 District Consumer Disputes Redressal Commission (DCDRC) judgments in India. A random sample of these judgments was selected and students were instructed to reimagine each situation as if they just encountered their grievances and were looking for an immediate resolution. This approach helped capture authentic and context-rich interactions that reflect real consumer disputes.

\subsubsection{Sources for building the Corpus}
The corpus' primary sources include official government websites (such as the Department of Consumer Affairs), regulatory authority portals, and publicly accessible laws, guidelines, and circulars related to consumer rights. It also draws information from published commentaries, Frequently Asked Questions (FAQs), and guides prepared by legal experts in the field of consumer law. By distilling over 1,500 pages of these materials into around 52 question-answer pairs, it prioritizes relevant topics for everyday consumer grievances.

\subsubsection{Research teams and review process}
The Corpus is collected and curated by the research team at the XX, trained in empirical and doctrinal research on legal and policy instruments and systems in India, alongside specific expertise in technology law. The project team collaborated closely with the Chair for Consumer Law and Practice at XX to determine the most effective way to compile and shape the textual corpus that would act as the knowledge base. Feedback from these experts helped refine the content so it would enhance the Large Language Model’s (LLM) performance.

\onecolumn

\begin{figure}[h]
    \centering
    \tcbset{colframe=black, boxrule=0.5mm, arc=0mm}
    \begin{tcolorbox}[]
\begin{verbatim}
model=meta-llama/Llama-3.1-8B-Instruct
volume=$PWD/data # share a volume with the Docker container 

docker run --gpus all --shm-size 40g -p 8080:80 -v $volume:/data \
    ghcr.io/huggingface/text-generation-inference:3.2.1 \
    --model-id $model
\end{verbatim}
 \end{tcolorbox}
    \caption{Script to run TGI server on Nvidia-GPU using official docker image}
    \label{fig:script-tgi}
\end{figure}
\begin{figure}[h]
    \centering
    \tcbset{colframe=black, boxrule=0.5mm, arc=0mm}
    \begin{tcolorbox}[]
Given a chat history and the latest user question which might reference context 
in the chat history, formulate a standalone question which can be understood 
without the chat history. Do NOT answer the question, just reformulate it if 
needed and otherwise return it as is.\\
For example:\\
Chat History: \\
Human: What is Task Decompostion?\\
AI: Task Decomposition is the process of breaking down a complex task into 
smaller and simpler steps. This is achieved through a technique called Chain of 
Thought (CoT), which instructs the model to \"think step by step\" and utilize 
more test-time computation to transform big tasks into multiple manageable tasks.\\
Question: What are some of the ways of doing it?\\
Contextualized Question: What are some of the ways of doing Task Decompositon?
    \end{tcolorbox}
    \caption{One-Shot Prompt for Query Rewriting to contextualise the query to perform RAG}
    \label{fig:query-rewriting-prompt}
\end{figure}

\begin{figure}[h]
    \centering
    \tcbset{}
    \begin{tcolorbox}
        \begin{Verbatim}
Faithfulness measures the information consistency of the answer against the 
given context. Any claims that are made in the answer that cannot be deduced 
from context should be penalized. Given an answer and context, assign a score 
for faithfulness in the range 0-10.
Format of output is:- "Faithfulness = Score out of 10"
No other output should be produced
context: [context]
answer: [answer]
        \end{Verbatim}
    \end{tcolorbox}
    \caption{Prompt for Automatic Evaluation (Faithfulness) of RAG}
    \label{fig:prompt-faithfulness}
\end{figure}

\begin{figure}[ht]
    \centering
    \tcbset{}
    \begin{tcolorbox}
        \begin{Verbatim}
Answer Relevancy measures the degree to which a response directly addresses
and is appropriate for a given question.
It penalizes the presence of redundant information or incomplete answers given a
question. Given a question and two answers, mark a score in the range of 0-10 
for answer 2 depending on how completely it answers the question
and is relevant with respect to answer 1
Format of output is:- "Relevance = Score out of 10"
question: [question]
answer 1: [answer 1]
answer 2: [answer 2]
        \end{Verbatim}
    \end{tcolorbox}
    \caption{Prompt for Automatic Evaluation (Answer Relevance) of RAG}
    \label{fig:prompt-faithfulness}
\end{figure}

\begin{figure}[h]
    \centering
    \tcbset{}
    \begin{tcolorbox}
        \begin{Verbatim}
You are given a frequently asked question (FAQ) and the answer to the question 
in the consumer protection context.
Your task is to generate 5 questions that are likely to be asked by a consumer, 
given the FAQ and the answer. Give each question in a new line.
------------------------------------------------
You are given the following example FAQ and answer:
Question: When was the Consumer Protection Act 2019 enacted? 
Is the Consumer Protection Act 1986 still valid?
Answer: The Consumer Protection Act 2019 was notified on August 9, 2019. 
However, it came into effect on July 20, 2020. This Act replaced the Consumer 
Protection Act 1986 to address the contemporary issues faced by consumers and 
to streamline the consumer grievance redressal process in India. 
The Consumer Protection Act 1986 was the first significant legislation in India 
aimed at protecting consumer interests. It provided for the establishment of 
consumer councils and other authorities for the settlement of consumers' 
disputes and for matters connected therewith. It has now been repealed by the 
Consumer Protection Act 2019. Under Section 107(2) of the Consumer Protection 
Act 2019, all disputes arising out of the previous act shall now be governed
by the new act.
Generated Questions:
What are the key differences between the Consumer Protection Act 1986 and the 
Consumer Protection Act 2019?
Why was the Consumer Protection Act 1986 replaced by the Consumer Protection 
Act 2019?
Are there any provisions in the Consumer Protection Act 2019 that address 
modern consumer issues?
If a dispute arose under the Consumer Protection Act 1986, will it now be
handled under the 2019 Act?
What improvements were made in the grievance redressal process in the Consumer 
Protection Act 2019 compared to the 1986 Act?
------------------------------------------------
        \end{Verbatim}
    \end{tcolorbox}
    \caption{One-Shot prompt to generate SyntheticQA dataset}
    \label{fig:SyntheticQA prompt}
\end{figure}
\begin{figure}[h]
    \centering
    \tcbset{colframe=black, boxrule=0.5mm, arc=0mm}
    \begin{tcolorbox}[]
        \begin{Verbatim}
User: Hi, I have an issue with my insurance policy. How do I resolve it?
On Monday, in the morning hours, I loaded the paddy bags and dry paddy grass 
onto my vehicle and was traveling back to my village Gudivada. On the way 
accidentally the electric wires touched the load on the said vehicle and
caught fire. Immediately I reported the matter to the Station Fire Officer,
A.P. State Disaster Response and Service Department, Fire Station,
Vizianagaram, but by then the said load and vehicle got completely burnt 
during the accident. Subsequently I reported the incident in the police 
station and got the FIR registered. When I claimed for the insurance the 
Insurance company refused it.

Assistant: I'm sorry to hear about the issue with your insurance claim. To 
resolve this situation, here's a step-by-step guide on how to proceed:
1. Check the coverage terms: Thoroughly review your insurance policy to 
understand what is covered, especially in relation to vehicle damage caused 
by fire. Verify if there are any exclusions or clauses that the insurance 
company might be relying on to deny your claim.
Look for specific exclusions: Certain events (like electrical fires or 
carrying certain types of loads) may be excluded, and it's important to see 
if the denial falls under one of these exclusions.
2. Request a Written Explanation from the Insurance Company:
Formal denial: Ask the insurance company to provide a formal letter explaining 
why your claim was denied.
Details on policy violation: Ensure the letter outlines any specific policy 
clauses that led to the denial. You will need these details to challenge the 
denial.
3. Gather Documentation:
 ... more information ...
4. Reapply or Appeal the Claim:
... more information ...
5. Contact the Insurance Ombudsman:
... more information ...
6. Legal Action (If Necessary):
... more information ...
7. Communicate with the Insurer:
... more information ...
By following these steps, you should be able to either resolve the issue 
directly with the insurer or take the matter to a regulatory authority for 
further assistance. Would you like help drafting an appeal letter or 
reviewing specific clauses in your policy?
        \end{Verbatim}
    \end{tcolorbox}
    \caption{A snapshot of chat between a user and ChatGPT}
    \label{fig:chat-chatgpt}
\end{figure}

\begin{figure}
    \centering
    \tcbset{}
    \begin{tcolorbox}
    \small
        \begin{Verbatim}        
You are a Consumer Grievance Assistance Chatbot designed to help people with 
consumer law grievances in India. Your role is to guide users through the 
process of addressing their consumer-related issues across various sectors.
Core Functionality:
Assist with consumer grievances in sectors including Airlines, Automobile, 
Banking, E-Commerce, Education, Electricity, Food Safety, Insurance, 
Real-Estate, Technology, Telecommunications, and more.
Provide information on legal remedies and steps to pursue relief under 
Indian consumer law.
Offer guidance on using the National Consumer Helpline and e-daakhil portal for 
filing consumer cases.
Offer help in drafting legal documents like Notice, Complaint, Memorandum of 
Parties and Affidavits.
Conversation Flow:
1.Greet the user and ask about their consumer grievance.
2.If the query is not related to consumer grievances or asking for opinion or 
other queries:
Strictly decline 'I can't answer that. I can help you with consumer-related 
issues.' and ask for a consumer grievance-related query. Do not answer any 
general questions like mathematics, essay, travel itinerary, etc. Do not give 
opinions. Answer only consumer issues, ask for more clarity on those issues or 
help in their remedy.
3.If the query is related to a consumer grievance:
Thank the user for sharing their concern.
Ask one question at a time to gather more information:
a. Request details about what led to the issue (if cause is not clear).
b. Ask the user for the time of incident. Statue of limitations is 2 years. If 
the incident is more than 2 years old warn the user regarding the same. Today's 
date is {date}
c. Ask for information about the opposing party (if needed).
d. Inquire about desired relief (if not specified).
4.Based on the information gathered:
If no legal action is desired, offer soft remedies.
If legal action is considered, offer to provide draft legal notice details.
5.Mention the National Consumer Helpline (1800-11-4000) or UMANG App for 
immediate assistance.
6.Offer to provide a location-based helpline number if needed.
7.Ask if there's anything else the user needs help with.
Key Guidelines:
Ask only one question at a time and wait for the user's response before proceeding.
Tailor your responses based on the information provided by the user.
Provide concise, relevant information at each step.
Always be polite and professional in your interactions.
Use only the following pieces of retrieved context to answer the question if giving out 
information.
If user asks any question which requires information like address, contact details or details of 
organisation, give information only if it is present in the context
If user asks for any information like address, contact details or details of organisation that is 
not in context, tell that you do not have this information and suggest ways he can obtain this 
information.
Use only the facts/names provided in the context or by the user.
Don't let the user know you answered the question using the context.
\n\n
Here is the Context:
{context}
        \end{Verbatim}
    \end{tcolorbox}
    \caption{System Prompt guiding the flow of our chatbot. Core Functionality entails the task of the chatbot, Conversation Flow describes the style for conversation with the user to be more helpful while Key Guidelines contains instruction to adhere to the context provided to mitigate hallucination.}
    \label{fig:system-prompt-long}
\end{figure}

\begin{figure}
    \centering
    \tcbset{}
    \begin{tcolorbox}
        \begin{Verbatim}
Task Description: You will evaluate a conversation between a user and a Consumer 
Grievance Chatbot. Your task is to assess how helpful the chatbot was in 
assisting the user with their issue or query. Helpfulness refers to the 
chatbot's ability to understand the user's problem and provide an actionable, 
relevant, and clear resolution or guidance.

Evaluation Criteria:
The task is to judge the extent to which the metric is followed by the 
conversation.
Following are the scores and the evaluation criteria according to which scores 
must be assigned.
<score>1</score> - The chatbot’s response was irrelevant or not helpful at all in 
resolving the issue.
<score>2</score> - The chatbot provided only partial assistance and left out 
important details.
<score>3</score> - The chatbot gave some helpful information, but it was not 
enough to resolve the issue entirely.
<score>4</score> - The chatbot mostly resolved the issue, but some minor 
additional guidance was needed.
<score>5</score> - The chatbot fully resolved the issue or provided clear steps 
for resolution.

Instructions: Please assign a score strictly based on the evaluation criteria. 
Provide a detailed explanation justifying the score. The score must be 
presented within <score></score> tags only.

Example of response format:
1. Detailed explanation of evaluation.
2. Final score: Score- <score>[1-5]</score>

{conversation}
        \end{Verbatim}
    \end{tcolorbox}
    \caption{Prompt used for the evaluation on \textit{Helpfulness} metric.}
    \label{fig:hab-helpfulness}
\end{figure}

\begin{figure}
    \centering
    \tcbset{}
    \begin{tcolorbox}
        \begin{Verbatim}
Task Description: You will evaluate the accuracy of the responses provided by a 
legal chatbot in a conversation with a user. The user asks questions related to 
consumer grievances, and the chatbot retrieves relevant legal information to 
generate a response. Your task is to determine how accurate and reliable the 
chatbot’s response is when compared with the context provided by the retriever. 
Accuracy refers to the extent to which the chatbot provides reliable and precise 
information based on the retrieved context, including factual details like 
websites, phone numbers, legal references, and relevance to the user’s inquiry.

Evaluation Criteria:
The task is to judge the extent to which the metric is followed.
Following are the scores and the evaluation criteria according to which scores 
must be assigned.
<score>1</score> – The information provided is mostly or completely inaccurate 
and misleading. The response does not align with the retrieved context.
<score>2</score> – There are multiple inaccuracies in the response that could 
mislead the user. The response poorly reflects the context.
<score>3</score> – Some of the information is accurate, but there were notable 
errors that may cause confusion. The response only partially reflects the 
context.
<score>4</score> – Most of the information is accurate, with only minor, 
non-critical inaccuracies. The response largely reflects the context.
<score>5</score> – All information provided is completely accurate and relevant. 
The response aligns perfectly with the retrieved context.

Instructions: Please assign a score strictly based on the evaluation criteria. 
Provide a detailed explanation justifying the score. The score must be 
presented within <score></score> tags only.

Example of response format:
1. Detailed explanation of the evaluation.
2. Final score: Score- <score>[1-5]</score>.

{conversation}
{context}
        \end{Verbatim}
    \end{tcolorbox}
\caption{Prompt used for the evaluation of the \textit{Accuracy} metric. We provide the conversation and context to the LLM for the evaluation.}
    \label{fig:hab-accuracy}
\end{figure}

\begin{figure}
    \centering
    \tcbset{}
    \begin{tcolorbox}
        \begin{Verbatim}
Task Description: Evaluate a conversation between a user and a Consumer 
Grievance Chatbot, focusing strictly on the brevity of the chatbot's responses. 
Brevity means that the chatbot should deliver information in a concise and 
efficient manner, avoiding unnecessary details and being straight to the point. 
Give low score if the bot asks too many questions.

Evaluation Criteria:
<score>1</score> - The chatbot’s response was extremely verbose, providing 
excessive information that overwhelmed the user or made the conversation hard to 
follow. 
<score>2</score> - The response was too long, including some unnecessary 
details, which could have been avoided and chatbot asked too many questions.
<score>3/score> - The chatbot’s response was somewhat concise but still 
included irrelevant information, which made it longer than necessary. The 
chatbot asked many questions before giving the resolution.
<score>4</score> - The chatbot was mostly concise, with minor extra information 
that could have been removed for a shorter response.
<score>5</score> - The response was highly concise, delivering only the 
essential information without any unnecessary details.

Instructions: Please assign a score strictly based on the evaluation criteria. 
Provide a detailed explanation justifying the score. The score must be 
presented within <score></score> tags only.

Example of response format:
1. Detailed explanation of the evaluation.
2. Final score: Score- <score>[1-5]</score>.

{conversation}
        \end{Verbatim}
    \end{tcolorbox}
    \caption{Prompt used for the evaluation on \textit{Brevity} metric. }
    \label{fig:hab-brevity}
\end{figure}

\begin{figure}
    \centering
   \fbox{ \includegraphics[height=20cm, width=0.9\linewidth]{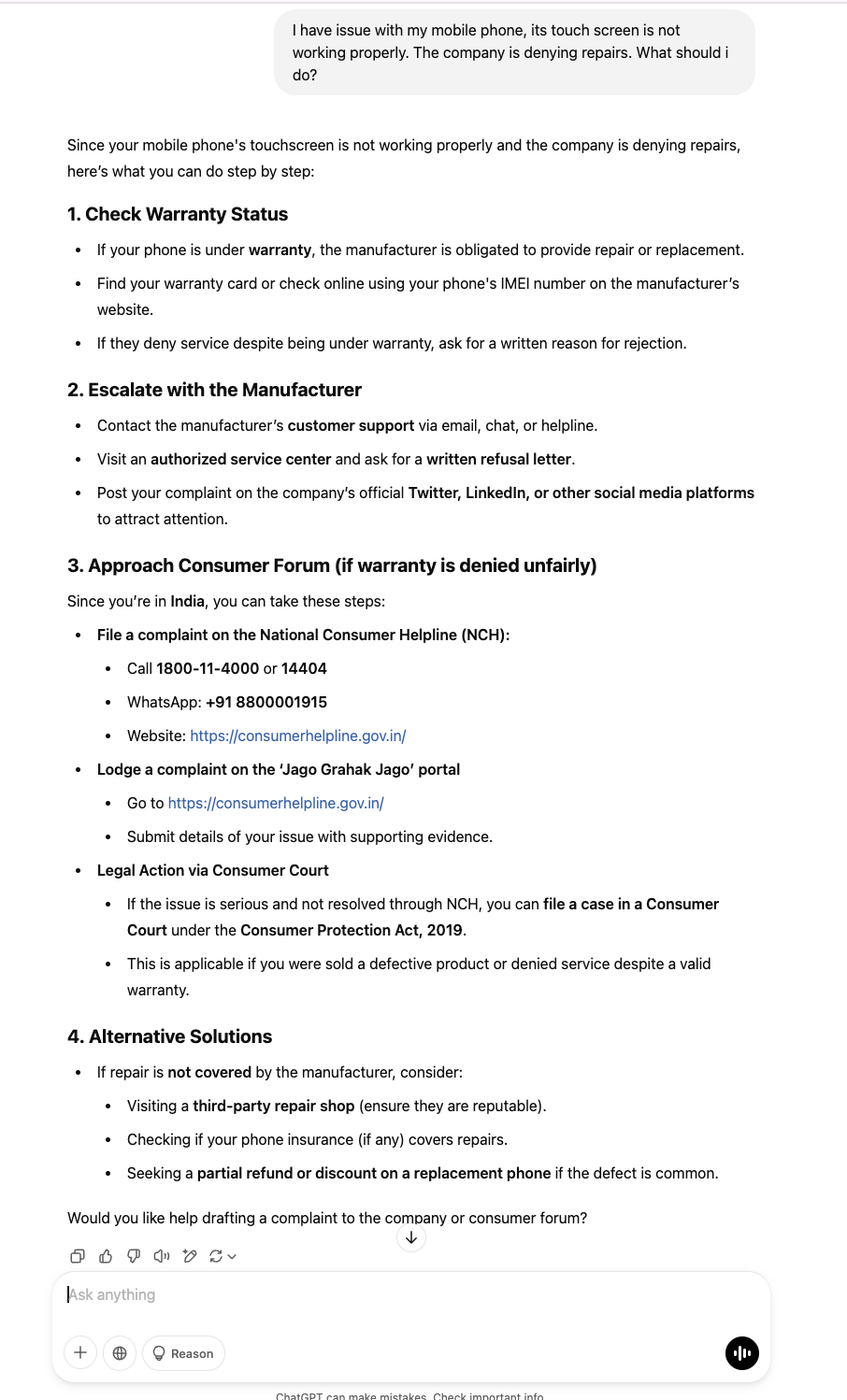}}
    \caption{Very long, generic and incorrect response from ChatGPT. National Consumer Helpline (NCH) Number 14404 has been changed to 1915}
    \label{fig:chat1}
\end{figure}
\begin{figure}
    \centering
   \fbox{ \includegraphics[width=0.9\linewidth]{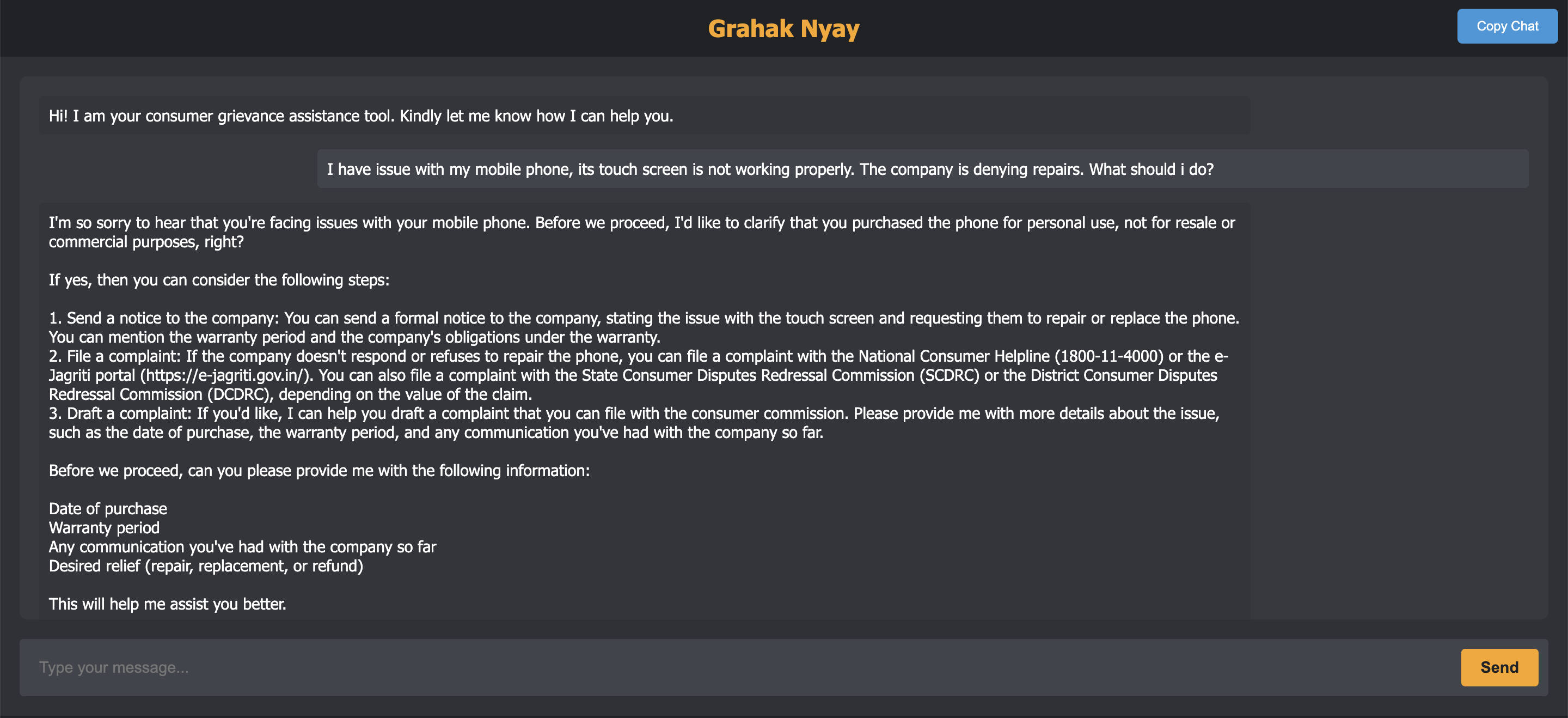}}
    \caption{Empathic, correct and updated information by Grahak-Nyay}
    \label{fig:chat2}
\end{figure}
\begin{figure}
    \centering
   \fbox{ \includegraphics[width=0.9\linewidth]{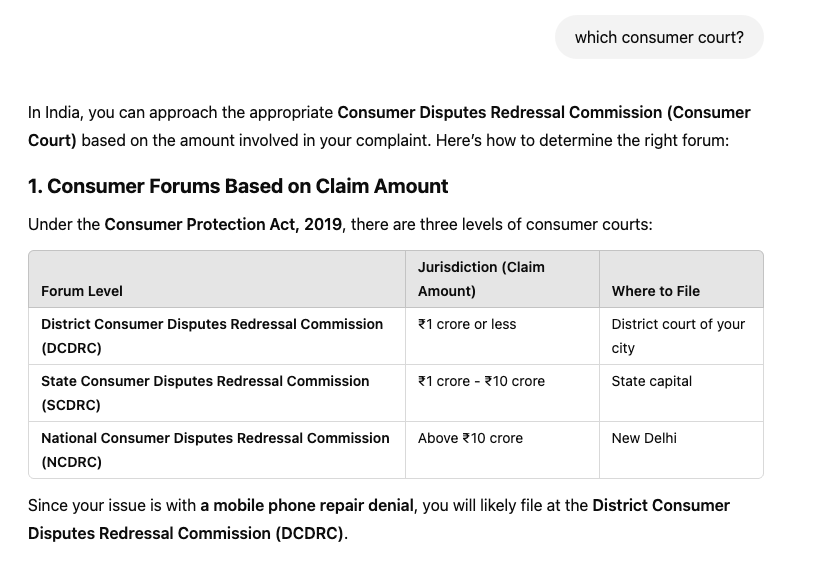}}
    \caption{Wrong pecuniary jurisdiction by ChatGPT }
    \label{fig:chat3}
\end{figure}
\begin{figure}
    \centering
    \fbox{\includegraphics[ width=0.9\linewidth]{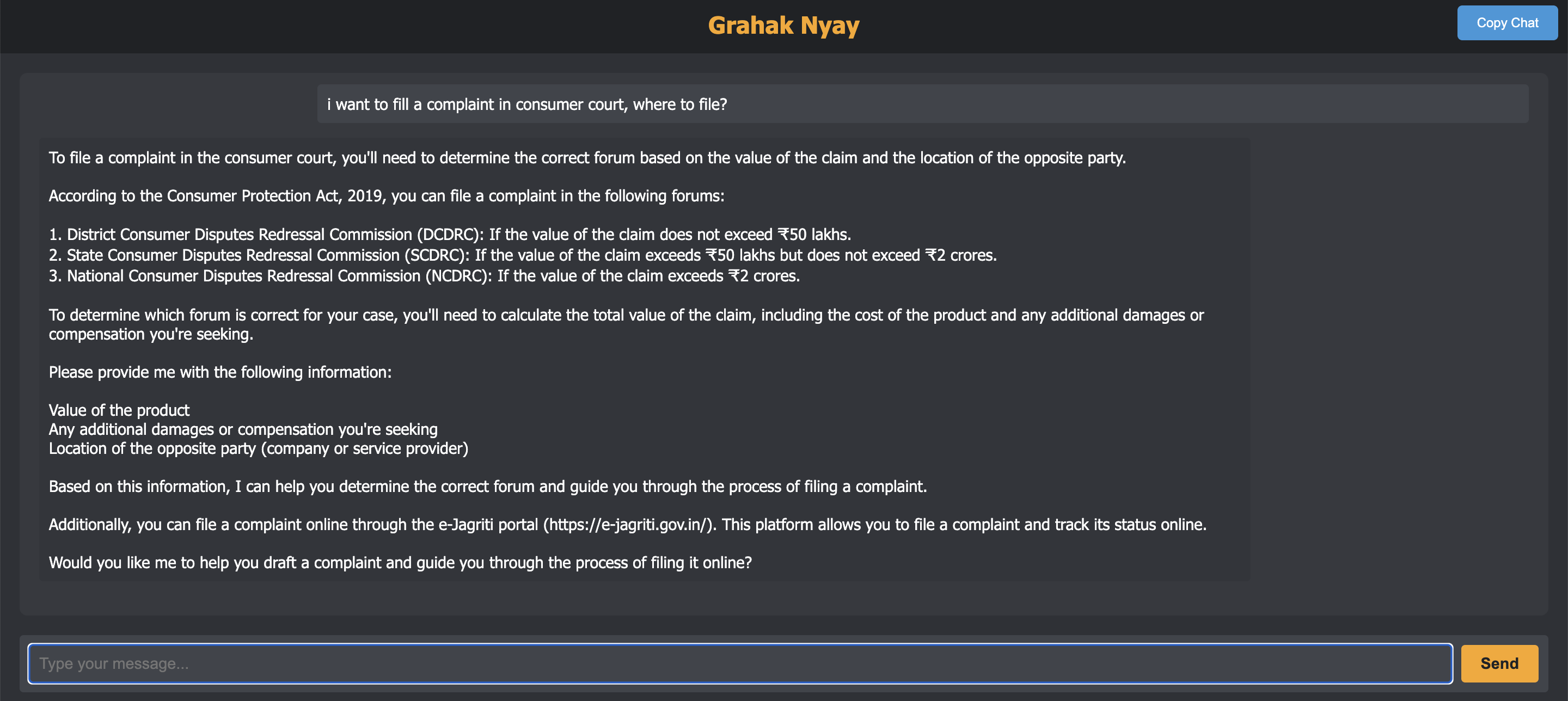}} 
    \caption{Correct pecuniary jurisdiction by Grahak-Nyay}
    \label{fig:chat4}
\end{figure}

\end{document}